# GENETIC ALGORITHM (GA) IN FEATURE SELECTION FOR CRF BASED MANIPURI MULTIWORD EXPRESSION (MWE) IDENTIFICATION

Kishorjit Nongmeikapam[1] and Sivaji Bandyopadhyay[2]

[1]Department of Computer Sc. and Engineering, Manipur Institute of Technology,
Manipur University, Imphal, India
`kishorjit.nongmeikapa@gmail.com`
[2]Department of Computer Sc. & Engg., Jadavpur University,
Jadavpur, Kolkata, India
`sivaji_cse_ju@yahoo.com`

## ABSTRACT

*This paper deals with the identification of Multiword Expressions (MWEs) in Manipuri, a highly agglutinative Indian Language. Manipuri is listed in the Eight Schedule of Indian Constitution. MWE plays an important role in the applications of Natural Language Processing(NLP) like Machine Translation, Part of Speech tagging, Information Retrieval, Question Answering etc. Feature selection is an important factor in the recognition of Manipuri MWEs using Conditional Random Field (CRF). The disadvantage of manual selection and choosing of the appropriate features for running CRF motivates us to think of Genetic Algorithm (GA). Using GA we are able to find the optimal features to run the CRF. We have tried with fifty generations in feature selection along with three fold cross validation as fitness function. This model demonstrated the Recall (**R**) of **64.08%**, Precision (**P**) of **86.84%** and F-measure (**F**) of **73.74%**, showing an improvement over the CRF based Manipuri MWE identification without GA application.*

## KEYWORDS

*CRF, MWE, Manipuri, GA, Features*

## 1. INTRODUCTION

This MWE is an important topic in the application of Natural Language Processing (NLP) like Part of Speech Tagging, Information Retrieval, Question Answering, Summarization, Machine Translation etc. The MWE is composed of an ordered group of words which can stand independently and carries a different meaning from its constituent words. For example in English: *'to and fro', 'bye bye', 'kick the bucket' etc*. MWEs include compounds (both word-compounds and phrasal compounds), fixed expressions and technical terms. A fixed expression MWE is one whose constituent words cannot be moved randomly or substituted without distorting the overall meaning or allowing a literal interpretation. Fixed expressions range from word-compounds and collocations to idioms. Some of the proverbs and quotations can also be considered as fixed expressions.

Manipuri is a highly agglutinative Indian Language spoken mainly in Manipur and in some other parts of the North Eastern India. Apart from India it is also spoken in some parts of Bangladesh and Myanmar. This language is a Tibeto-Burman type of language. Manipuri language is a Scheduled Language of Indian Constitution.

Manipuri uses two scripts; one is the borrowed Bengali Script while the other one is its original Meitei Mayek (Script). The Manipuri with Bengali Script is adopted in this work since the corpus of Meitei Mayek Manipuri is so far hard to collect. The development of an automatic MWE system requires either a comprehensive set of linguistically motivated rules or a large amount of annotated corpora in order to achieve reasonable performance.

Different attempts of using CRF are found in order to find Multiword Expression for other language and even for Manipuri but the main handicap is in the selection of the features for CRF. That is the reason a hybrid model of identifying MWE using CRF as in [1] and GA [2] is designed. The feature selection of CRF model is not an easy task. So far the CRF based MWE feature selection is manual or is simply a hit and trial method. A new approach to tackle this issue is the implementation of GA. A simple model of Genetic Algorithm (GA) is adopted in order to choose the features as in [3].

The paper is organized with related works of MWE in Manipuri and other languages in Section 2 followed by the motivation and challenges of Manipuri in Section 3. Section 4 and Section 5 present the basic concepts of CRF and Genetic Algorithm, Section 6 mentions a simple stemming rule for Manipuri. Section 7 lists all the features for running the CRF and the hybrid model of Manipuri MWE identification is discussed. Section 8 describes the experiments and the evaluation. The conclusion and the future works road map are drawn in Section 9.

## 2. RELATED WORKS

The works on MWEs can be seen in [4]-[7]. For Indian languages also works are being done to identify the MWEs [8]-[12]. The published works on identifications of NER and MWEs in Manipuri are also found. For the NER, works are reported in [13]-[15]. Manipuri MWE works are reported in [1] and reduplicated MWEs in [16] and [17]. The identification of Manipuri MWEs is quite difficult since the root words in Manipuri are only noun and verb, words of different part of speech are derived from them. The stemming work in Manipuri can be found in [18] and [19].

## 3. CHALLENGES AND MOTIVATION

### 3.1. Challenges in identification of MWE NEs in Indian languages

The notable work of [13] gives us the idea about the challenges and difficulties in working with Manipuri like other Indian Languages:

1. Unlike English and most of the European languages, Manipuri lacks capitalization information, which plays a very important role in identifying Name Entities (NEs). So it is a problem in the identification of Named Entities which may be MWEs.

2. A lot of NEs in Manipuri can appear in the dictionary with some other specific meanings. So sometimes it creates confusion between MWE NE and normal words.

3. Manipuri is a highly inflectional language providing one of the richest and most challenging sets of linguistic and statistical features resulting in long and complex wordforms.

4. Manipuri is a relatively free word order language. Thus NEs can appear in subject and object positions making the identification of MWE NEs task more difficult.

5. Manipuri is a resource-constrained language. Annotated corpus, name dictionaries, sophisticated morphological analyzers, POS taggers etc. are not yet available.

With the above mention challenges one need to carefully adopt a method so that optimal output is generated. These challenges also motivate us in the identification of MWE.

### 3.2. The Agglutinative Nature

The most important and challenging thing about Manipuri is the word structure which is highly agglutinative. The affixes are bundled up one after another, specially the suffixes. Altogether 72 (seventy two) affixes are listed in Manipuri out of which 11 (eleven) are prefixes and 61 (sixty one) are suffixes. Table 1 shows the 10 prefixes. The prefix ম (mə) is used both as formative and pronomial prefix but it is included only once in the list. Similarly, Table 2 lists 55 (fifty five) suffixes as some of the suffixes are used with different forms of usage such as ঙম (gum) which is used as particle as well as proposal negative, দা (də) as particle as well as locative and না (nə) as nominative, adverbial, instrumental or reciprocal.

To prove the point that Manipuri is highly agglutinative let us site an example word: "*পুসিনহনজারমগাদাবানিদাকো*" (pusinhənjərəmgədəbənidəko), which means "(I wish I) myself would have caused to carry in (the article)". Here there are 10 (ten) suffixes being used in a verbal root, they are "*pu*" is the verbal root which means "to carry", "*sin*"(in or inside), "*hən*" (causative), "*jə*" (reflexive), "*rəm*" (perfective), "*gə*" (associative), "*də*" (particle), "*bə*" (infinitive), "*ni*" (copula), "*də*" (particle) and "*ko*" (endearment or wish).

Table 1. Prefixes

| Prefixes used in Manipuri |
|---|
| অ, ই, ই, থু, চা, ত, থ, ন, ম and শে |

Table 2. Suffixes

| Suffixes used in Manipuri |
|---|
| কন, কুম, কো, খরে, খৎ, খাই, খি, খোয়, গা, গনি, গী, ঙম, ঙৈ, চা, চো, থ, থৎ, থেক, থোক, দা, দি, দুনা, দে, না, নতে, নি, নিং, নু, নে, পী, ফাও, বা, বু, মক, মল, মিন, মুক, লে, লা, লক, ল্ল, লি, লী, লু, লূ লে, লো, লোয়, শনু, শি, শিং, শিন, শু, হৎ and হন |

## 4. CONCEPTS OF CONDITIONAL RANDOM FIELD (CRF)

The concept of Conditional Random Field as in [20] has been developed in order to calculate the conditional probabilities of values on other designated input nodes of undirected graphical models. CRF encodes a conditional probability distribution with a given set of features. It is an unsupervised approach where the system learns by some training and can be used for testing other texts.

The conditional probability of a state sequence $Y=(y_1, y_2,..y_T)$ given an observation sequence $X=(x_1, x_2,..x_T)$ is calculated as :

$$P(Y|X) = \frac{1}{Z_X} \exp(\sum_{t=1}^{T}\sum_{k} \lambda_k f_k(y_{t-1}, y_t, X, t)) \quad ---(1)$$

where, $f_k(y_{t-1}, y_t, X, t)$ is a feature function whose weight $\lambda_k$ is a learnt weight associated with $f_k$ and to be learned via training. The values of the feature functions may range between $-\infty \ldots +\infty$, but typically they are binary. $Z_X$ is the normalization factor:

$$Z_X = \sum_{y} \exp \sum_{t=1}^{T}\sum_{k} \lambda_k f_k(y_{t-1}, y_t, X, t) \quad ---(2)$$

which is calculated in order to make the probability of all state sequences sum to 1. This is calculated as in the Hidden Markov Model (HMM) and can be obtained efficiently by dynamic programming. Since CRF defines the conditional probability P(Y|X), the appropriate objective for parameter learning is to maximize the conditional likelihood of the state sequence or training data.

$$\sum_{i=1}^{N} \log P(y^i \mid x^i) \qquad \text{---(3)}$$

where, $\{(x^i, y^i)\}$ is the labeled training data.

Gaussian prior on the $\lambda$'s is used to regularize the training (i.e., smoothing). If $\lambda \sim N(0,\rho^2)$, the objective function becomes,

$$\sum_{i=1}^{N} \log P(y^i \mid x^i) - \sum_{k} \frac{\lambda_i^2}{2\rho^2} \qquad \text{---(4)}$$

The objective function is concave, so the $\lambda$'s have a unique set of optimal values.

## 5. CONCEPTS OF GENETIC ALGORITHM (GA)

Genetic Algorithm is a probabilistic search method that has been developed by John Holland in 1975. Genetic Algorithm is the mimic of the real world for natural selection, mutation, crossover and production of the new offspring. Genetic Algorithm helps us in natural evolution. The implementation of GA in Natural Language Processing is a new direction and interest of research. Basic steps or algorithm in GA are as follows.

**Algorithm:** *Genetic Algorithm*

**Step 1:** Initialize a population of chromosomes.

**Step 2:** Evaluate each chromosome in the population or chromosome pool.

**Step 3:** Create offspring or new chromosomes by mutation and crossover from the pool.

**Step 4:** Evaluate the new chromosomes by a fitness test and insert them in the population.

**Step 5:** Check for stopping criteria, if satisfies return the best chromosome else continue from step 3.

**Step 6:** End

Figure1 explains the above algorithm as a flow chart. The Genetic Algorithm have five basic components, they are:

1. Chromosome representations for the feasible solutions to the optimization problem.
2. Initial population of the feasible solutions.
3. A fitness function that evaluate each solution.
4. A genetic operator that produce a new population from the existing population.

Control parameter like population size, probability of generic operators, number of generations etc.

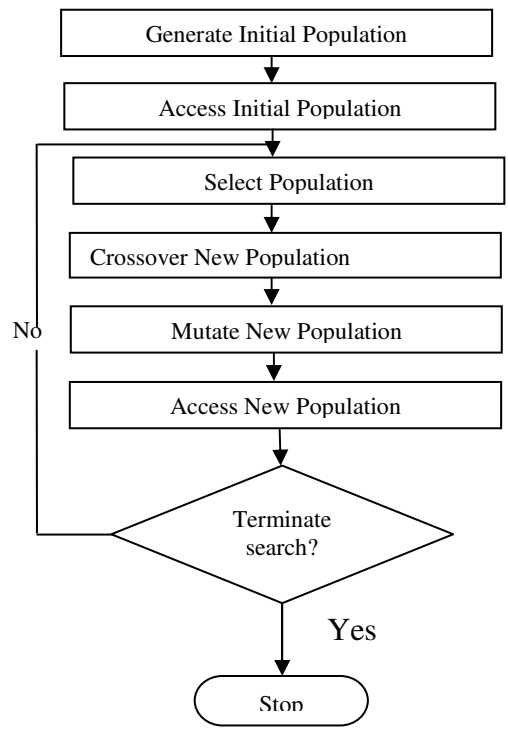

Figure 1. Flow chart of Genetic Algorithm.

## 6. MANIPURI STEMMING ALGORITHM

Manipuri being a highly agglutinative language the best way of stemming is that the Manipuri words are stemmed by stripping the suffixes in an iterative manner as mentioned in [18]. In order to stem a word an iterative method of stripping is done by using the acceptable list of prefixes (11 numbers) and suffixes (61 numbers) as mentioned in the Table 1 and Table 2 above.

### 6.1 Algorithm

As mention in [18] the design of this stemmer mainly consist of four algorithms the first one is to read the prefixes, the second one is to read the suffixes, the third one is to identify the stem word removing the prefixes and the last algorithm is to identify the stem word removing the suffixes.

Two file, *prefixes_list* and *suffixes_list* are created for prefixes and suffixes of Manipuri. In order to test the system another testing file, *test_file* is used. (We use Java programming language thus String are considered as object).

We remove the prefixes and suffixes in an iterative approach as shown in the algorithm 3 and algorithm 4 until all the affixes are removed. The stem word is stored in *stemwrd.*

**Algorithm1:** `read_prefixes()`

```
1. Repeat 2 to 4 until all prefixes (p_i) are read from
   prefixes_list
2. Read a prefix p_i
3. p_array[i]=p_i
4. p_wrd_count =i++;
5. exit
```

**Algorithm2:** `read_suffixes()`

1. Repeat 2 to 4 until all suffixes ($s_i$) are read from `suffixes_list`
2. Read a suffix $s_i$
3. `s_array[i]=`$s_i$
4. `s_wrd_count=i++;`
5. `exit`

**Algorithm3:** `Stem_removing_prefixes(p_array, p_wrd_count)`

1. Repeat 2 to 16 for every word ($w_i$) are read from the `test_file`
2. `String stemwrd=" ";`
3. `for(int j=0;j<p_wrd_count;j++)`
4. `{`
5. `if(`$w_i$`.startsWith(p_array[j]))`
6. `  {`
7. `  stemwrd=`$w_i$`.substring(`$w_i$`.length()-((`$w_i$`.length()-((p_array[j].toString()).length()))),`$w_i$`.length());`
8. `  `$w_i$`=stemwrd;`
9. `  j=-1;`
10. `  }`
11. `else`
12. `  {`
13. `   stemwrd=`$w_i$`;`
14. `  }`
15. `}`
16. `write stemwrd;`
17. `exit;`

**Algorithm4:** `Stem_removing_suffixes(s_array,s_wrd_count)`

1. Repeat 2 to 16 for every word ($w_i$) are read from the `test_file`
2. `String stemwrd=" ";`
3. `for(int j=0;j<s_wrd_count;j++)`
4. `  {`
5. `  if(`$w_i$`.endsWith(s_array[j]))`
6. `   {`
7. `  stemwrd=`$w_i$`.substring(0,`$w_i$`.indexOf(s_array[j]));`
8. `  `$w_i$`=stemwrd;`
9. `   j=-1;`
10. `  }`
11. `  else`
12. `  {`
13. `   stemwrd=`$w_i$`;`
14. `  }`
15. `}`
16. `write stemwrd;`
17. `exit`

# 7. INTEGRATED MODEL OF IDENTIFYING MANIPURI MWE

An integrated model is adopted here which can perform CRF based MWE identification as in [1] but changes are made with the feature list and feature selection. The feature selection is applied with the concept of Genetic Algorithm. The advantage here is the way of selecting the features.

## 7.1 The CRF Model of Identifying MWE Manipuri

For the current work C++ based CRF++ 0.53 package[1] which is readily available as open source for segmenting or labeling sequential data is used. The CRF model for identification of Manipuri MWE (Figure 2) consists of mainly data training and data testing. The following subsection will brief about each step of this CRF model we have used.

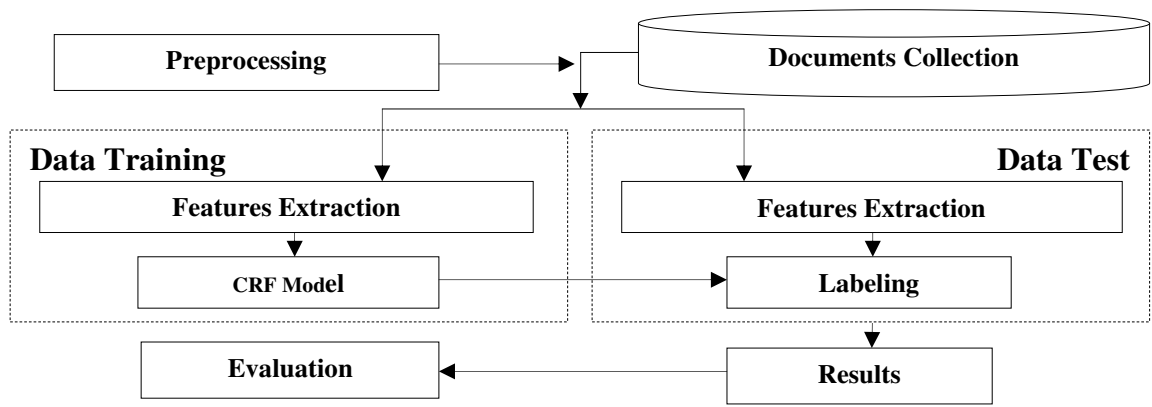

Figure 2.  CRF Model of Identifying MWE

## 7.2 The Feature List

In order to get the best result, a careful listing of features is important in CRF. The various candidate features are listed as follows:

**F= {$W_{i-m}$, …, $W_{i-1}$, $W_i$, $W_{i+1}$,... , $W_{i-n}$, $SW_{i-m}$, …, $SW_{i-1}$, $SW_i$, $SW_{i+1}$,... , $SW_{i-n}$ , Acceptable suffixes present in the word, Binary notation if suffix is present, Number of acceptable suffixes, Acceptable prefixes present in the word, Binary notation if prefix is present, Binary Notation of general salutations in previous words, Binary Notation of general follow up words of Multiword Name Entities, Digit feature, Word length, Word frequency and its surrounding word frequency, Surrounding POS tag}**

The details of the set of features that have been applied for identification of MWE in Manipuri are as follows:

  **Current word and surrounding words:** The current word and surrounding words are the focal point of MWE so selecting the current word and surrounding words as a feature is important.

  **Surrounding Stem words as feature:** Stemming is done as mentioned in Section 6 and the preceding and following stem words of a particular word with the stem of the current word are used as features since the preceding and following words influence the present word in case of MWE.

---

[1] http://crfpp.sourceforge.net/

**Acceptable suffixes:** 61 suffixes have been manually identified in Manipuri and the list of suffixes is used as one feature. As mentioned with an example in Section 3, suffixes are appended one after another and the maximum number of appended suffixes can be ten. So taking such cases into account, ten columns separated by space for each word to store every suffix present in the word. A "0" notation is being used in those columns when the word consists of less or no acceptable suffixes.

**Acceptable prefixes as feature:** 11 prefixes have been manually identified in Manipuri and the list of prefixes is used as a feature. For every word the prefix is identified and a column is created mentioning the prefix if it is present, otherwise the "0" notation is used.

**Binary notation for suffix(es) present:** The suffixes play an important role in Manipuri since it is a highly agglutinative language. For every word if suffix(es) is/are present during stemming a binary notation '1' is used, otherwise a '0' is stored.

**Number of acceptable suffixes as feature:** For every word the number of suffixes is identified during stemming, if any and the number of suffixes is used as a feature.

**Binary notation for prefix(es) present:** The prefixes play an important role in Manipuri since it is a highly agglutinative language. For every word if prefix(es) is/are present during stemming a binary notation '1' is used, otherwise a '0' is stored.

**Binary Notation of general salutations/preceding word of Name Entity:** Name Entities are generally MWEs. In order to identify the NE which are MWE, salutations like Mr., Miss, Mrs, Shri, Lt., Captain, Rs., St., Date etc that precede the Name Entity are considered as a feature for the MWE. A binary notation of '1' if used, else a '0' is used.

**Binary notation of general follow up words of Name Entity:** As mentioned above, Name Entities are generally MWEs. The following word of the current word can also be considered as a feature since a name may have ended up with clan name or surname or words like 'organization', 'Lup' etc for organization, words like 'Leikai', 'City' etc for places and so on. A binary notation of '1' if used else a '0' is used.

**Digit features:** Date, currency, weight, time etc are generally digits. Thus the digit feature is an important feature. A binary notation of '1' is used if the word consists of a digit else a '0' is used.

**Length of the word:** Length of the word is set to 1 if it is greater than 3. Otherwise, it is set to 0. Very short words are rarely MWE.

**Word and surrounding word frequency:** A range of frequencies for words in the training corpus are identified: those words with frequency <100 occurrences are set to the value 0, those words which occurs >=100 times but less than 400 times are set to 1. The word and its surrounding words frequency are considered as one feature since MWEs are rare in occurrence compared to those of determiners, conjunctions and pronouns.

**Surrounding POS tag:** The POS of the surrounding words are considered as an important feature since the POS of the surrounding words influence the MWE.

### 7.3 Preprocessing

A Manipuri text document is used as an input file. The training and test files consist of multiple tokens. In addition, each token consists of multiple (but fixed number) columns where the columns are used by a template file. The template file gives the complete idea about the feature selection. Each token must be represented in one line, with the columns separated by white spaces (spaces or tabular characters). A sequence of tokens becomes a **sentence**. Before undergoing training and testing in the CRF, the input document is converted into a multiple token file with fixed columns and the template file allows the feature combination and selection.

An example sentence formation of few words in the model for feeding in the CRF tool is as follows:

অদুগা অদু গা 0 0 0 0 0 0 0 0 1 1 0 0 0 0 1 1 CC O O
ইংলিস ইংলিস 0 0 0 0 0 0 0 0 0 0 ই 1 0 0 0 1 0 NNP O O
স্কুলশিং স্কুল শিং 0 0 0 0 0 0 0 0 1 1 0 0 0 0 1 0 NN O O
অমদি অমদি 0 0 0 0 0 0 0 0 0 0 0 0 0 0 0 1 5 CC O O
যুনিভর্সিটিশিংদা যুনিভর্সিটি দা শিং 0 0 0 0 0 0 0 1 2 0 0 0 0 0 1 0 NLOC O O
লুরক্লবা লুরক্ল বা 0 0 0 0 0 0 0 0 1 1 0 0 0 0 1 0 VN O O
মতাঙদা মতাঙ দা 0 0 0 0 0 0 0 0 1 1 ম 1 0 0 1 1 RB O O
মমালোন্দা মমালোন্ দা 0 0 0 0 0 0 0 0 1 1 ম 1 0 0 1 0 NN O O
তাক্লরিবা তাক্লরি বা 0 0 0 0 0 0 0 0 1 1 ত 1 0 0 1 0 VN O O
অসিনা অ না সি 0 0 0 0 0 0 0 1 2 0 0 0 0 0 1 3 PR O O

Figure 3. Example Sample Sentence in the Training and the Testing File

An example of the template file which consists of feature details for two example stem words before the word, two stem words after the word, current word, the suffixes (upto a maximum of 10 suffixes), binary notation if suffix is present, number of suffixes, the prefix, binary notation of prefix is present, binary notation if digit is present, binary notation if general list of salutation or preceding word is present, binary notation if general list of follow up word is present, frequency of the word, word length, POS of the current word, POS of the prior two word, POS of the following two word details is as follows:

```
# Unigram
U00:%x[-2,1]
U01:%x[-1,1]
U02:%x[0,1]
U03:%x[1,1]
 U04:%x[2,1]
U05:%x[-1,1]
U06:%x[0,0]
U10:%x[0,2]
U11:%x[0,3]
U12:%x[0,4]
U13:%x[0,5]
U14:%x[0,6]
U15:%x[0,7]
U16:%x[0,8]
U17:%x[0,9]
U18:%x[0,10]
U20:%x[0,11]
U21:%x[0,12]
U22:%x[0,13]
U23:%x[0,14]
U24:%x[0,15]
U25:%x[0,16]
U26:%x[0,17]
U27:%x[0,18]
U28:%x[0,19]
U29:%x[0,20]
```

```
    U30:%x[0,21]
    U31:%x[-1,21]
    U32:%x[-2,21]
    U33:%x[0,21]
    U34:%x[1,21]
    U35:%x[2,21]
# Bigram
```

Figure 4. Example template file

To run the CRF generally two standard files of multiple tokens with fixed columns are created: one for training and another one for testing. In the training file the last column is manually tagged with all those identified MWEs using the tags of B-MWE and I-MWE for the beginning and rest of the MWE respectively else 'O' for those which are not MWE. In the test file we can either use the same tagging for comparisons and evaluation or only 'O' for all the tokens regardless of whether it is MWE or not.

### 7.4 The Features in Chromosome representation and its selection procedure

As mentioned in Section 5 the Chromosome pool or population is developed. Each chromosome consists of genes, which is binary valued. When the gene value is '1' then the feature is selected and when it is '0' the feature is not selected. Figure 4 demonstrates the feature representation as a chromosome.

Feature

| $F_1$ | $F_2$ | $F_3$ | ... | $F_{n-2}$ | $F_{n-1}$ | $F_n$ |
|---|---|---|---|---|---|---|

Chromosome

| 1 | 0 | 1 | ... | 0 | 1 | 1 |
|---|---|---|---|---|---|---|

Feature subset = {F1, F3,..., Fn-1, Fn}

Figure 5. Example Feature encoding using GA

From the Chromosome pool or initial population, first of all initial selection of chromosome is done. A randomly selected crossover point is marked and **crossover** is executed. During crossover, the sub parts of genes are exchanged from the two chromosomes at the crossover point.

The **mutation** is also done in random so that the chromosomes are not repeated. The objective of mutation is restoring the lost and exploring variety of data. The bit value is changed at a randomly selected point in the process of mutation.

Three fold cross validation technique is used as a **Fitness function**. By three fold cross validation we mean dividing the corpus into 3 nearly equal parts for doing 3-fold cross-validation (use 2 parts for training and the remaining part for testing and do this 3 times with a different part for testing each time).

After fitness test the chromosomes which are fit are placed in the pool and the rest of the chromosomes are deleted to create space for those fit ones.

### 7.5 Model File after training

In order to obtain a model file we train the CRF using the training file. This model file is a ready-made file by the CRF tool for use in the testing process. In other words the model file is the learnt file after the training of CRF. We do not need to use the template file and training file again since the model file consists of the detail information of the template file and training file

### 7.6 Testing

The test file is the test data where sequential tags of the MWEs will be assigned else 'O' is assigned for those words which are not MWEs. This file has to be created in the same format as that of the training file, i.e., fixed number of columns with the same fields as that of training file.

The output of the testing process is a new file with an extra column which is tagged with B-MWE and I-MWE for the beginning and rest of the MWE respectively else a 'O' is tagged for those which are not MWEs

## 8. EXPERIMENT AND EVALUATION

Manipuri corpus are collected and filtered to rectify the spelling and syntax of a sentence by a linguist expert from Linguistic Department, Manipur University. In the corpus some words are written in English, such words are rewritten into Manipuri in order to avoid confusion or error in the output. The corpus we have collected includes 45,000 tokens which are of Gold standard.

Evaluation is done with the parameters of Recall, Precision and F-score as follows:

Recall,

$$R = \frac{No\ of\ correct\ ans\ given\ by\ the\ system}{No\ of\ correct\ ans\ in\ the\ text}$$

Precision,

$$P = \frac{No\ of\ correct\ ans\ given\ by\ the\ system}{No\ of\ ans\ given\ by\ the\ system}$$

F-score,

$$F = \frac{(\beta^2 + 1)\ PR}{\beta^2 P + R}$$

Where $\beta$ is one, precision and recall are given equal weight.

A number of problems have been faced while doing the experiment due to typical nature of the Manipuri language. In Manipuri, word category is not so distinct. The verbs are also under bound category. Another problem is to classify basic root forms according to the word class. Although the distinction between the noun class and verb classes is relatively clear; the distinction between nouns and adjectives is often vague. Distinction between a noun and an adverb becomes unclear because structurally a word may be a noun but contextually it is adverb. Further a part of root may also be a prefix, which leads to wrong tagging. The verb morphology is more complex than that of noun. Sometimes two words get fused to form a complete word.

### 8.1 Experiment for selection of best feature

As mention earlier in 3-fold cross validation, we divide the corpus into 3 nearly equal parts for doing 3-fold cross-validation (use 2 parts for training and the remaining part for testing). A total of 45,000 words are divided into 3 parts, each of 15000 words.

The features are selected using the GA and experiments are performed in order to identify the best feature. The best features are those which gave the maximum recognition of MWE in a given text. The experiment is conducted with 50 generations.

In each run of the CRF tool the feature template are changed according to the chromosome selected. Table 3 shows the result in terms of Recall (**R**), Precision (**P**) and F-measure (**F**).

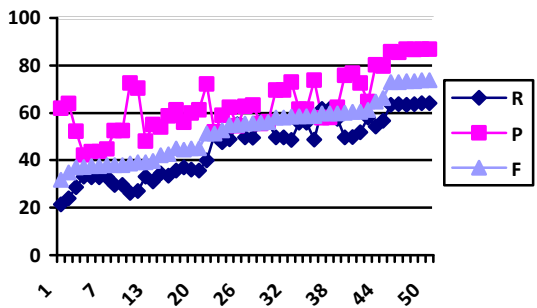

Figure 6. Chart of the performance in 50 generations

The System stops running when the F-Score shows no change in the l score, i.e., when the output is exhausted.

## 8.2 Evaluation and best Feature

The earlier model of CRF based identification of MWE in Manipuri as mentioned in [1] uses the following feature list with manual selection:

**F= {$W_{i-m}$, …, $W_{i-1}$, $W_i$, $W_{i+1}$,... , $W_{i-n}$ , |prefix|<=n, |suffix|<=n, Surrounding POS tag, word length, word frequency, acceptable prefix, acceptable suffix}**

The list consists of surrounding words, prefixes, suffixes, surrounding POS, word length, word frequency, acceptable prefix and acceptable suffix. Improvement has been observed using reduplicated MWE as additional feature.

The model which has been adopted here has a different list and the best feature is chosen after the best performance, i.e., when saturated output is generated. The best result is considered when the system output is saturated, i.e, when there is no change in the output. This happens with the following feature:

**F= {$W_{i-2}$, $W_{i-1}$, $W_i$, $SW_{i-1}$, $SW_i$, Upto 5 acceptable suffixes present in the word, Binary notation if suffix is present, Number of acceptable suffixes, Binary notation if prefix is present, Binary Notation of general salutations in previous words, Binary Notation of general follow up words of Mutiword Name Entities, Digit feature, Word length, Word frequency of previous two words, Current word POS tag, Following two words POS tag}**

The best feature set in the model gives the Recall (**R**) of **64.08%**, Precision (**P**) of **86.84%** and F-measure (**F**) of **73.74%**.

The earlier model in [1] reports that the CRF based system shows a recall of 60.39%, precision of 85.53% and F-measure of 70.83%. It is also reported that with reduplicated MWEs as one feature it makes an improvement in implementation of CRF and thus the new improved recall as reported earlier is 62.24%, precision is 86.06% and F-measure is 72.24%.

The model adopted in this paper when compared with the exclusive CRF based MWE identification as in [1] shows an improvement of **2.91%** in **F-Score** also when compare with the earlier model which has improvement done with reduplicated MWE shows a better **F-score** of **1.5%**.

In the case of **Recall, 3.69%** is recorded comparing with the earlier model and **1.84%** improvement is found comparing with the previous improved model with reduplicated MWE as added feature.

An improvement of **1.31%** in the **Precision** too can be noticed and **0.78%** when compared with the improved CRF using reduplicated MWE.

The main advantage in this model is in the selection of the feature where it is done with the application of GA

## 9. CONCLUSION

So far different approaches for identification of MWE in Manipuri are found but implementation of Genetic Algorithm (GA) was never attempted. This model has come up with the successful implementation of GA in feature selection of CRF for the first time in Manipuri language. Implementation can also be tried for the other unsupervised methods with features. With this method, the burden of manual feature selection is reduced. This method can be tried for other Indian agglutinative languages.

Using GA for feature selection we are able to find the optimal features to run the CRF. We have tried with fifty generations in feature selection along with three fold cross validation as fitness function. This model demonstrated the Recall (**R**) of **64.08%**, Precision (**P**) of **86.84%** and F-measure (**F**) of **73.74%,** showing an improvement over the CRF based Manipuri MWE identification without GA application.

**Authors**

Short Biography

Kishorjit Nongmeikapam is working as Asst. Professor at Department of Computer Science and Engineering, MIT, Manipur University. He has completed his BE from PSG college of Tech., Coimbatore and has completed his ME from Jadavpur University. He is presently doing research in the area of Multiword Expression and its applications. He has so far published 14 papers and presently handling a Transliteration project funded by DST, Govt. of Manipur, India.

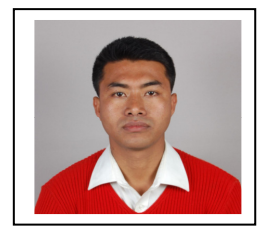

Professor Sivaji Bandyopadhyay is working as a Professor since 2001 in the Computer Science and Engineering Department at Jadavpur University, Kolkata, India. His research interests include machine translation, sentiment analysis, textual entailment, question answering systems and information retrieval among others. He is currently supervising six national and international level projects in various areas of language technology. He has published a large number of journal and conference publications.

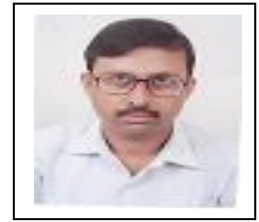